\documentclass{article}

\usepackage{natbib}
\usepackage{amsmath}
\usepackage{amsfonts}
\usepackage{graphicx,psfrag,epsf}
\usepackage{enumerate}
\usepackage{float}
\usepackage{bm}
\usepackage{subcaption}
\usepackage{url} 
\usepackage{authblk}
\usepackage[left=1.5cm, right=2cm]{geometry}
\newtheorem{remark}{Remark}
\newtheorem{theorem}{Theorem}
\newcommand{\Lin}{\text{L}_{\text{in}}}
\newcommand{\Lout}{\text{L}_{\text{out}}}
\newcommand{\gammaT}{\mathbf{\gamma}_{0}}

\newcommand{\jacobT}{\nabla_{\mathbf{\gamma}}\mathcal{L}(\mathbf{\gammaT})}

\DeclareMathOperator*{\argmax}{argmax}
\DeclareMathOperator*{\argmin}{argmin}

\begin{document}

\title{Representation Learning of Dynamic Networks}

\author[1]{Haixu Wang}
\author[2]{Jiguo Cao}
\author[3]{Jian Pei}

\affil[1]{Department of Mathematics and Statistics, University of Calgary, Calgary, AB, Canada \authorcr \texttt{haixu.wang@ucalgary.ca}}
\affil[2]{Department of Statistics and Actuarial Science, Simon Fraser University, Burnaby, BC, Canada \authorcr \texttt{jiguo\_cao@sfu.ca}}
\affil[3]{Department of Computer Science, Duke University, Durham, NC, USA \authorcr \texttt{j.pei@duke.edu}}

\date{}


\maketitle

\newpage
\begin{abstract}
This study presents a novel representation learning model tailored for dynamic networks, which describes the continuously evolving relationships among individuals within a population. The problem is encapsulated in the dimension reduction topic of functional data analysis. With dynamic networks represented as matrix-valued functions, our objective is to map this functional data into a set of vector-valued functions in a lower-dimensional learning space. This space, defined as a metric functional space, allows for the calculation of norms and inner products. By constructing this learning space, we address (i) attribute learning, (ii) community detection, and (iii) link prediction and recovery of individual nodes in the dynamic network. Our model also accommodates asymmetric low-dimensional representations, enabling the separate study of nodes' regulatory and receiving roles. Crucially, the learning method accounts for the time-dependency of networks, ensuring that representations are continuous over time. The functional learning space we define naturally spans the time frame of the dynamic networks, facilitating both the inference of network links at specific time points and the reconstruction of the entire network structure without direct observation. We validated our approach through simulation studies and real-world applications. In simulations, we compared our method’s link prediction performance to existing approaches under various data corruption scenarios. For real-world applications, we examined a dynamic social network replicated across six ant populations, demonstrating that our low-dimensional learning space effectively captures interactions, roles of individual ants, and the social evolution of the network. Our findings align with existing knowledge of ant colony behavior. In summary, we propose a statistical representation learning model for dynamic networks that strikes a balance between interpretability and learning capacity, offering comprehensive insights into network dynamics.
\end{abstract}


\newpage

\section{Introduction}
\label{sec:introduction}
\begin{figure}[htbp]
\centering
\includegraphics[width=\textwidth]{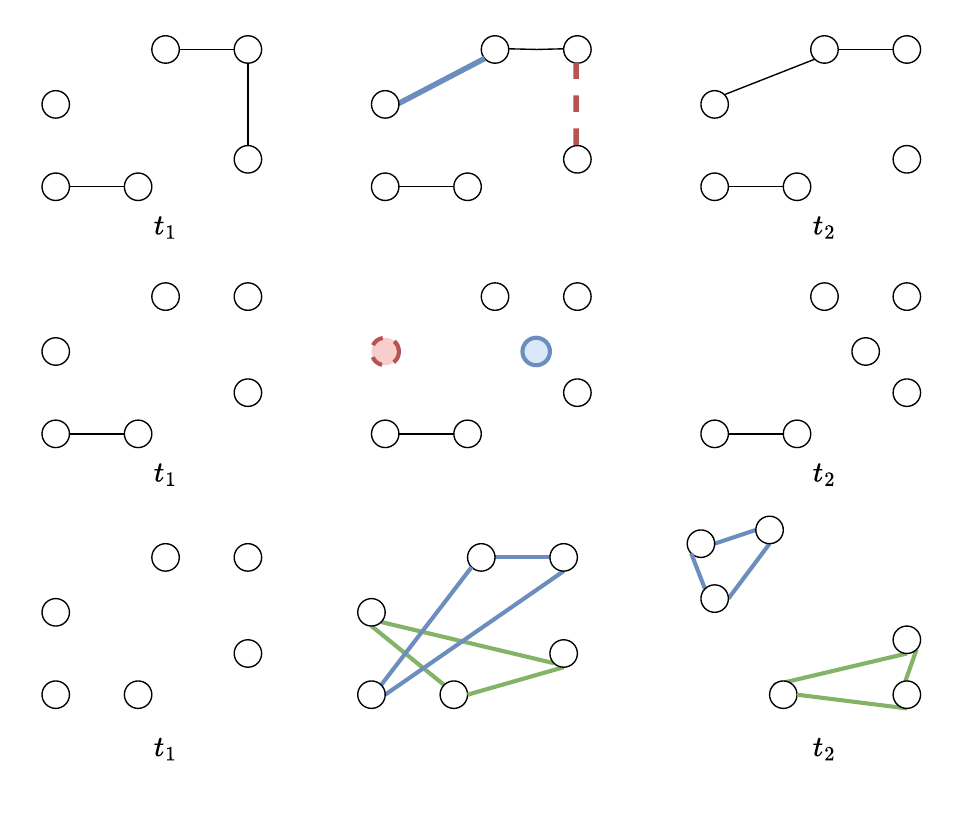}
\caption{\label{fig:networkpreview} Three common changes that may happen in dynamic networks. The first row illustrates the insertion (blue) and deletion (red) of links from time point $t_{1}$ to $t_{2}$. The second row depicts the addition (blue) and deletion (red) of nodes in the dynamic network from time point $t_{1}$ to $t_{2}$. The third row presents how communities are constructed through linking individual nodes of the dynamic network.}
\end{figure}

This work establishes a representation learning model for dynamic networks which describes the continuously changing relationship between individuals in a population. Some examples of dynamic networks include social networks, real-time trading networks, and gene regulatory networks. Figure~\ref{fig:networkpreview} demonstrates an example of a dynamic network where connections between nodes change over time. The changes include adding/removing links, inserting/deleting nodes, and constructing communities through links.\newline

\noindent Representation learning became a necessity in building statistical models to address the ever-increasing complexity and scale of real data. A representation learning model does not only simplify real data but often reveals underlying information that is more useful than the actual observations. In comparison to traditional data types in statistics, networks are unique and difficult to analyze because individual traits are encoded in interactions with others rather than direct measurements of individuals. We can refer to observed interactions as intrinsic information of a network, and it is possible to observe extrinsic information for each individual node in the network. Sometimes, we need to use network information to predict other variables, but directly combining a node's interactions (as in integer-valued vectors) with other measured values is not optimal for subsequent prediction tasks. Through representation learning of networks, we could represent each node in the network as a coordinate in a representation space, where the dimensionality of such space is much lower than the number of nodes in the network. Furthermore, the representation space is a real-valued vector space, hence node's representation (coordinates) can be easily concatenated with other information observed for the node. We can see that the representation learning model is a bridge between network information and other data types. 

Representation learning of a static network is to represent each node in the network by a real-valued vector. The dimension of the representation space is much less than the number of nodes in the network. It is equivalent to representing an adjacency matrix with a smaller matrix where we will reduce the number of columns compared to the dimension of the original adjacency matrix. The learning procedure is deeply rooted in the topic of matrix factorization. Two major approaches are principal component analysis and non-negative matrix factorization (NMF). The former is discussed in \cite{PCAMF} and \cite{eigen_graphembedding} which is to project nodes' connection vectors onto a subset of eigenvectors. The projections will serve as the low-dimensional representations of individual nodes. Non-negative matrix factorization \citep{NFM_lee} is another way of decomposing the adjacency matrix. For instance, \cite{NFM_networkembedding} has demonstrated the application of NMF on network embedding while preserving the community structures of nodes. Another type of approach is to use random walks to explore a network, where transitions between steps in a walk are determined by the conditional probability of a connection between a pair of nodes. Methodologies are developed through \cite{LINE, deepwalk, randomwalk_graphembedding}, and \cite{node2vec}. Later, \cite{surveyNE} provides a survey on the current methodologies of network embedding. Existing works on representation learning of a dynamic (time-dependent) network are mainly about extending the current network embedding methods. For example, both \cite{CTDNE} and \cite{dynnode2vec} allow random walks to travel through time instead of being constrained on a single time snapshot of the dynamic network.

Representation learning of dynamic networks is much more difficult to achieve. Dynamic networks add another time-dependent structure to analyze, so we have to consider how to represent nodes in a continuous space of time to understand how nodes interact with each other in a network. Dynamic networks are difficult to analyze and even more so for building prediction models compared to static networks or common multivariate data. 

To the best of our knowledge, this is the first attempt to design a statistical representation learning model to sufficiently compress information encoded in a dynamic network. The core objective of building this representation learning model is to establish the representation space with its associated mapping process. More importantly, we need the representation space to have a support of time that addresses the evolution of dynamic networks. Instead of stringing separate representation spaces at different time points together, it is more suitable to consider a representation space that is naturally defined over the support of time. In fact, the desired representation space can be represented by a functional space.  

This motivates us to use functional data analysis \citep{ramsayFDA,Ferraty06,Kokoszka12,Eubank15} to frame the problem in learning representation functions for a matrix-valued function. Observed network structures will be the adjacency matrix at time points as well as discrete evaluations of the matrix-valued function. Given that the representation space is a functional space, there are a few conditions for such representation space to be useful: (i) continuity of representation in time, (ii) equipped with norm and inner product, and (iii) preserving network (observed/hidden and static/dynamic) topological features. The first condition is easily met because the representation space will be a function (of time) space. The second condition can be satisfied by entailing the representation space in a Hilbert space which has a complete metric. The last condition will be enforced throughout the construction of the representation space through penalization as well as introducing asymmetry in the representation space. The details will be later outlined in Section~\ref{sec:methods}.

Representation learning is rooted in functional data analysis through the framework of basis expansion. The gist of the framework is to represent a random function, an infinite-dimensional object, by a linear combination of a set of basis functions. Variations of this framework are to use different bases, e.g., splines and eigenfunctions. Expressing functions in terms of splines has been widely used in functional data analysis. Uses of splines are discussed in \cite{Wahba_1990}, \cite{guidetosplines} and \cite{ramsayFDA}, and the asymptotics have been studied in both \cite{localasymptotics} and \cite{AsymptoticofPsplines}. Another technique for analyzing functional data is functional principal component analysis (FPCA), which uses a set of eigenfunctions as a basis to represent any random individual functional data \citep{PCAformultivariate, supervised_fpca, iFPCA}. We can also use basis expansion and FPCA in the context of general linear models as discussed in \cite{functionalGLM}. The theories and asymptotics of FPCA are studied in \cite{fpca_asmp_1,fpca_asmp_2,fpca_asmp_3}.


We propose a representation learning model to embed a dynamic network into a lower-dimensional space that satisfies the following conditions: (i) continuity in time, (ii) equipped with norm and inner product, and (iii) preserving network topological features. To the best of our knowledge, this is the first statistical model that tackles the representation learning of a dynamic network. The current literature in either statistics or network analysis does not consider all three conditions as previously discussed. 

In short, we need to design a mapping function to represent a matrix-valued function, a time-varying adjacency matrix, with a set of vector-valued functions. Each vector-valued function represents an individual of the network across time. We will refer to the representation space as a functional space $\mathcal{F}$, and the mapping function will be denoted as $F$. With the help of functional data analysis, the latter functional space $\mathcal{F}$ can be defined as an inner-product space that induces the norm hence metrics for the space. At each time point, the connections of an individual node (a vector with size as the number of nodes in a network) are represented by a real-valued vector with much lower dimensions. Hence, we can see that the proposed model automatically meets the first two conditions and address the third condition through the learning process. 

The model also has the following benefits. First, the functional space $\mathcal{F}$ is defined over the support of time and naturally allows the representation of individual nodes to change continuously along the support. This approach is different and innovative compared to the existing works on learning dynamic networks, where the latter usually separates dynamic networks into snapshots at different time points. The representation space at different time points does not share information or preserve the continuity or underlying evolution of the network structure. 

Second, we will define a functional space with an inner product that can be easily calculated and interpreted. The inner product will facilitate the calculation of norms and distances between elements in the space. That is, we can calculate the overall distance between any pair of nodes in the dynamic network. If the situation demands, we can further reduce the functional space $\mathcal{F}$ to a standard vector space. Any prediction modeling can be done based on the subsequent vector space. In the meantime, the functional space $\mathcal{F}$ allows us to calculate the distance between any pair of nodes in the learning space at a specific time point. We can use these distances to describe the network structure as well as connections among individual nodes at any point in the time interval. 

The third benefit of the proposed model is that the representation space $\mathcal{F}$ is constructed by pooling network information together across all observed time points. This preserves the time-varying network topologies. In contrast to the existing works on learning dynamic networks, our proposed method does not separate dynamic networks into snapshots at different time points. The representation learning model should be built at once rather than on a sequence of discrete points. Our approach allows us to recover network structures where we do not have observations. In the meantime, we obtain time-dependent representations that change continuously in time and are more interpretable on the time-dependent network structures. For example, if we observe two nodes in the dynamic network to have their first connection, it is reasonable to assume there has been some momentum built in the representation space. Equivalently, the distance between these two nodes should decrease upon a connection has been observed, If the connection between a pair of nodes diminishes over time, then the distance between these two nodes should increase or remain at a degree that is bigger than the nodes to which they are connected. The same reasoning can be applied to model the deletion and addition of nodes. 

The last but not least benefit of our proposed method is that we will enable the representation space to preserve the community and group structures in the observed dynamic networks. During the model estimation step, we will force similar nodes in the network to reside closer in the representation space $\mathcal{F}$. If a group of nodes is much more connected or shares a similar connection in the network (statically or dynamically), we should observe them in a tighter cluster in $\mathcal{F}$. This will not only allow us to understand pairwise connections but also uncover the relationship between groups or sub-populations in a larger population. 

The manuscript is organized as follows. Section~(\ref{sec:methods}) outlines the representation learning model and estimation procedure. Section~(\ref{sec:theory}) discusses the limiting behaviors and distribution of the embedding function. Section~(\ref{sec:simulation}) includes simulation studies to assess the reliability and prediction performances of the proposed method in comparison to some existing works. Section~(\ref{sec:applications}) presents a real data application on a dynamic network that describes the social and physical interaction of a population of ants \citep{antnet}. Section~(\ref{sec:discussion}) concludes the paper with discussions and possible extensions of the current work.   

\section{Method}
\label{sec:methods}
Consider a time-varying network $G(t) = \{V(t), E(t)\}$ where the network structure changes over time $t \in [0,T]$. The evolution of the network $G(t)$ is described through that of the node population $V(t)$ and corresponding edge list $E(t)$. The node list $V(t)$ contains the individual nodes that are presenting in the network at time $t$ where $|V(t)|$ corresponds to the number of nodes. The time-dependent change of $V(t)$ could be deletions and additions of individual nodes. The dynamics of $E(t)$ reflect the simultaneous connections between individual nodes belonging to $V(t)$ at $t$. In this work, we assume that the dynamic network $G(t)$ is a directed and unweighted network across the time interval. The network $G(t)$ is observed in the corresponding adjacency matrix $A(t)$ at a sequence of time point $t_{1} = 0, ..., t_{n} = T$. The entry $A_{jk}(t)$ for $j,k \in V(t)$ indicates whether a connection has been observed from $j$-th node to $k$-th node. For simplicity, we will consider $M = \text{max}(|V(t)|)$ to be the maximum number of nodes after moving across the time interval.

Our aim of representation learning of a dynamic network $G(t)$ is to consider a mapping procedure $F$ such that 
\begin{equation}
 F: \mathbb{I}^{M \times M \times [0, T]} \rightarrow \mathbb{R}^{M \times R \times  [0, T]}
    \label{eq:representationmapping}
\end{equation}
with $R$ being much smaller than $M$. The representation mapping $F$ considers three learning tasks. The first task is to decompose the $M$ by $M$ adjacency matrix containing the observed network connections. The mapping finds an $R$-dimensional representation for each node in the network which summarizes connections observed in the adjacency matrix. The second task is to extend the first task over the entire support $[0, T]$. Hence, each node in $V$ (the supremum set on the possible nodes) has an $R$-dimensional continuous-valued vector where the probability of connection between each pair of nodes is proportional to their representations across all time points. The second task, in the sense of functional data analysis, is to find a representation mapping to convert an $N \times N$ dimensional functional data into a set of $R$-dimensional embedding functions for each node in $V$. The third task is about preserving the community structures of $G(t)$ through the mapping $F$. We will introduce community centers of nodes in the representation space and enforce the nodes to be closer to the centers if they share similar connections in the network. Preserving the topological features will be achieved through the penalization of individual representations. The rest of this section will present how the proposed representation learning methodology achieves these tasks.

For the first task, we consider that the node embeddings (real-valued representations) are asymmetric in the sense that each node in the network has two embeddings: (i) outgoing and (ii) receiving one. The asymmetry in node embeddings allows us to accommodate directed networks and study how each node behaves in both giving and receiving connections. Without loss of generality or affecting the modeling, we will use $V$ to represent the supremum set of nodes that will be observed throughout $[0, T]$. If a node is added to the network at a later time, we consider it to be present in the network from the beginning without any connections. We define the $j$-th node, for $j = 1,..., M$, in $V$ to have two embedding components $\bm{\alpha}_{j}(t)$ and $\bm{\beta}_{j}$ which reflect the asymmetry in network connections. The embedding function $\bm{\alpha}_{j}(t)$ carries the regulatory or outgoing information for the $j$-th node on others, whereas the embedding $\bm{\beta}_{j}$ represents how a node should receive information from others. At a time point $t$, $\{\bm{\alpha}_{j}(t)\}_{j=1}^{M}$ and $\{\bm{\beta}_{j} \}_{j=1}^{M}$ help us to decompose and analyze network connections observed in the corresponding adjacency matrix $\bm{A}(t)$. Having separate embedding components (in and out) grants the flexibility to explore the roles of individual nodes, e.g., receivers, senders, community hubs, or more. In the meantime, the two components are still in the same vector space $\mathbb{R}^{R}$ which maintains the inner product $\langle \bm{\alpha}_{j}(t), \bm{\beta}_{k} \rangle = \sum_{r=1}^{R}\bm{\alpha}_{jr}(t)\beta_{kr}$ for any $t \in [0, T]$ and $ j,k = 1, ..., M$. Given the two embedding components, it allows us to represent the network structure at any given time $t$ in the $R$-dimensional vector space.

The second task is to compress the dynamic information of a network and properly define the embedding functions $\mathbf{\alpha}_{j}(t)$'s along its space $\mathcal{F}$. To simplify modeling and computational complexity as well as preserve identifiability, we reduce the receiving embedding function $\bm{\beta}_{j}(t)$ to time-invariant vectors $\bm{\beta}_{j}$. Hence, we can focus on the $R$-dimensional functional space $\mathcal{F}$ in which $\{\bm{\alpha}_{j}(t)\}$'s are defined. As stated as one of the conditions, we want the $\mathcal{F}$ to have norms and distances calculated based on inner products. That is, for any $\bm{u}, \bm{v} \in \mathcal{F}$, there exists an inner product~(\ref{eq:func_innerproduct}) which induces a norm and metric for $\mathcal{F}$
\begin{equation}
  \langle \bm{u}(t), \bm{v}(t) \rangle_{\mathcal{F}} = \sum_{r=1}^{R}\int_{0}^{T}u_{r}(t)v_{r}(t)dt.
  \label{eq:func_innerproduct}
\end{equation}
We will use $\langle \cdot, \cdot \rangle_{\mathcal{F}}$ to serve as a basis for further calculations used for node clustering, classification, and more.

The third task of preserving community and network topology. In addition to embedding a time-varying network, we are often tasked to utilize the embedding space to perform other statistical modeling problems, e.g., node clustering, classification, regression, or more. In this work, we want to focus on node clustering which helps us to identify communities as well as roles of nodes in the network. If we observe communities in the observed network or have prior information from experts, the embedding space and node embeddings should preserve such information. In addition, the embedding space should be a better space to discover any hidden network topological structures given less dimensionality. For the remainder of the section, we will frame these aspects in the problem of node clustering in the embedding spaces. Since the node embeddings are asymmetric, we need to consider clustering in both $\bm{\alpha}_{j}(t)$'s and $\bm{\beta}_{j}$'s. We can assume that the cluster labels (either $\bm{\alpha}$ or $\bm{\beta}$) of a node are constant throughout the time during model estimation. On the other hand, it is also possible to allow the cluster labels to change over time given the embedding functions $\bm{\alpha}_{j}(t)$'s are continuous in time. After obtaining both embedding components of all nodes, we can run the clustering algorithm dynamically over time to get a more detailed clustering of nodes. We will demonstrate this analysis in the application section.

For clustering the nodes in the embedding space, we used the distance-based K-mean clustering method. Let $\text{L}_{\text{out}}$ and $\text{L}_{\text{in}}$ be the number of clusters in $\bm{\alpha}(t)$ and $\bm{\beta}$, then clustering the embedding components is done through the following minimization~(\ref{eq:kmean_out}) and ~(\ref{eq:kmean_in})
\begin{equation}
  \argmin_{\{\bm{\omega}_{l}(t)\}_{l=1}^{\text{L}_{\text{out}}}}\sum_{j=1}^{M}\argmin_{l \in 1, ..., \text{L}_{\text{out}}}||\bm{\alpha}_{j}(t) - \bm{\omega}_{l}(t)||_{\mathcal{F}}^{2}
  \label{eq:kmean_out}
\end{equation}
and
\begin{equation}
  \argmin_{\{\bm{\zeta}_{l}\}_{l=1}^{\text{L}_{\text{in}}}}\sum_{j=1}^{M}\argmin_{l \in 1, ..., \text{L}_{\text{in}}}||\bm{\beta}_{j} - \bm{\zeta}_{l}||^{2}
  \label{eq:kmean_in}
\end{equation}
which are objective functions of two K-mean clustering problems on $\bm{\alpha}_{j}(t)$'s and $\bm{\beta}_{j}$'s given cluster centers $\bm{\omega}_{l}(t)$'s and $\bm{\zeta}_{l}$'s.

\subsection{Model Specification}

We have introduced the two embedding components $\bm{\alpha}_{j}(t)$ and $\bm{\beta}_{j}$. The former requires functional data analysis treatment which defines the following form~(\ref{eq:bspline_alpha})
\begin{equation}
  \alpha_{ir}(t) = \sum_{d=1}^{D}\gamma_{ird}\phi_{d}(t)
  \label{eq:bspline_alpha}
\end{equation}
in the $r$-th dimension for $r = 1,..., R$. The set $\{\phi_{d}(t)\}_{d=1}^{D}$ are known functions and serve as a basis for representing any unknown function $\alpha_{ir}(t)$. Furthermore, we will assume that all $R$ dimensions of $\bm{\alpha}(t)$ will be using the same set of $\{\phi_{d}(t)\}_{d=1}^{D}$. That is, $\alpha_{r}(t)$ and $\alpha_{r^{\prime}}(t)$ are different in the coefficients $\bm{\gamma}_{r} = (\gamma_{r1}, ..., \gamma_{rD})^{T}$ and $\bm{\gamma}_{r^{\prime}} = (\gamma_{r^{\prime}1}, ..., \gamma_{r^{\prime}D})^{T}$.

The minimization of~(\ref{eq:kmean_out}) can be reduced to a simpler problem, i.e., from minimization in a functional space to a vector space of coefficients. Let the $l$-th cluster center $\bm{\omega}_{l}(t)$ to use the same set of basis functions as $\bm{\alpha}(t)$ in~(\ref{eq:outclustercenter}) such that 
\begin{equation}
  \omega_{lr}(t) = \sum_{d=1}^{D}\theta_{lrd}\phi_{d}(t)
  \label{eq:outclustercenter}
\end{equation}
and the norm $||\cdot||_{\mathcal{F}}^{2}$ in the functional space $\mathcal{F}$ is defined to be as~(\ref{eq:functionalnorm})
\begin{equation}
  ||\bm{\alpha}_{j}(t) - \bm{\omega}_{l}(t)||_{\mathcal{F}}^{2} = \sum_{r=1}^{R}\int_{\tau} (\alpha_{ir}(t) - \omega_{lr}(t))^{2}dt = \sum_{r=1}^{R}(\bm{\gamma}_{ir} - \bm{\theta}_{lr})^{T}\bm{\Phi}(\bm{\gamma}_{ir} - \bm{\theta}_{lr})
  \label{eq:functionalnorm}
\end{equation}
where $\bm{\Phi}$ is an $D$ by $D$ matrix with with $d,d^{\prime}$-th entry being $\int_{0}^{T}\phi_{d}\phi_{d^{\prime}}dt$.

Taking an element $A_{jk}(t)$, for $j, k = 1,..., M$, in the time-dependent adjacency matrix $\bm{A}(t)$, we consider it to be a binary-valued function which a probability function
\begin{equation}
  P(A_{jk}(t) = 1) = p_{jk}(t) = \frac{ \exp(\langle \bm{\alpha}_{j}(t), \bm{\beta}_{k} \rangle) }{1 + \exp(\langle \bm{\alpha}_{j}(t), \bm{\beta}_{k} \rangle)} .
  \label{eq:probmodel}
\end{equation}
We can see that eq.~(\ref{eq:probmodel}) corresponds to the functional logistic regression model which has a linear component:
\begin{equation}
  \eta_{jk}(t) = \text{log}(\frac{p_{jk}(t)}{1-p_{jk}(t)}) = \langle \bm{\alpha}_{j}(t), \bm{\beta}_{k}\rangle .
  \label{eq:logitmodel}
\end{equation}

Given that dynamic network has been observed on the time points $t_{1}, ..., t_{n}$, the embedding components are estimated through the maximization of the log-likelihood as follows:
\begin{align}
  \mathcal{L} &= \sum_{i=1}^{n}\sum_{j,k = 1}^{M} A_{jk}(t_{i}) \log(p_{jk}(t_{i})) + (1-A_{jk}(t_{i}))\log(1 - p_{jk}(t_{i})) \nonumber \\
  &= \sum_{i=1}^{n}\sum_{j,k = 1}^{M} A_{jk}(t_{i})\eta_{jk}(t_{i}) + \log(1 - p_{jk}(t_{i})),
  \label{eq:loglikelihood}
\end{align}
and 
\begin{equation*}
  \{\hat{\bm{\alpha}}_{j}(t),\hat{\bm{\beta}}_{k}\}_{j,k=1}^{M} = \argmax_{\{\gamma_{j}, \bm{\beta}_{k}\}_{j,k}}\mathcal{L}
\end{equation*}
with $\gamma_{j} = (\bm{\gamma}^{T}_{j1}, ..., \bm{\gamma}_{jR}^{T})^{T}$. 

The continuity in the basis functions $\phi_{d}(t)$'s transforms to that in the embedding function $\alpha_{r}(t)$'s. This is beneficial for learning the time-varying dynamics of a network. For example, the connection of a node to others should change continuously over time rather than experience dramatic isolation or saturation across time points, unless there are unknown network internal mechanisms to detect or outside factors that are alternating the network. Therefore, having continuous embedding function $\bm{\alpha}(t)$ allows us to understand the natural underlying evolution of a time-varying network. We can also study the dynamics of the network by looking at the derivatives of $\bm{\alpha}(t)$'s. 

Another benefit of representing $\bm{\alpha}(t)$ as in eq.~(\ref{eq:bspline_alpha}) is to (i) predict links of the network at unobserved time points and (ii) discover unobserved connections within time points and (iii) potential connections before or after deletion/addition of nodes into the network. For the first case, we can recover the network connections at any unobserved $t$ by evaluating $\bm{\alpha}(t)$'s at such $t$. For the second case, if a pair of nodes, $j$, and $k$-th nodes, go from unconnected to connected between time points $t_{1}$ to $t_{2}$, then there should be evidence in the embedding space through the change in the inner product between $\eta_{jk}(t_1)$ and $\eta_{jk}(t_{2})$ or $\eta_{kj}(t_1)$ and $\eta_{kj}(t_{2})$. Such evidence will be reflected in the dynamic representation $\bm{\alpha}(t)$. Having embedding functions will also help with tracing additions and deletions of nodes in the dynamic network. If a node is added to the network with connections to the other existing nodes, we should be able to see evidence building up in the embedding space. On the other hand, if a node is deleted from the network, it should have a lingering effect on other nodes where its embeddings should not diminish suddenly to others through the inner product.

\subsection{Model Estimation}

The proposed representation learning model for a dynamic network is now composed of $\bm{\alpha}_{j}(t)$'s, $\bm{\beta}_{j}$'s, $\bm{\omega}_{l}(t)$'s, and $\bm{\zeta}_{l}$'s, which are (i) the out-going node embedding functions, (ii) receiving embedding vector, (iii) cluster centers of $\bm{\alpha}(t)$, and (iv) cluster centers of $\bm{\beta}$ respectively. To estimate these components, we will start with the following objective function~(\ref{eq:totalobj})
\begin{equation}
  \mathcal{L} - \lambda_{\alpha 0}\min_{\{\bm{\omega}_{l}(t)\}_{l=1}^{\text{L}_{\text{out}}}}\sum_{j=1}^{M}\min_{l \in 1, ..., \text{L}_{\text{out}}}||\bm{\alpha}_{j}(t) - \bm{\omega}_{l}(t)||_{\mathcal{F}}^{2} - \lambda_{\beta 0}\min_{\{\bm{\zeta}_{l}\}_{l=1}^{\text{L}_{\text{in}}}}\sum_{j=1}^{M}\min_{l \in 1, ..., \text{L}_{\text{out}}}||\bm{\beta}_{j} - \bm{\zeta}_{l}||^{2}
  \label{eq:totalobj}
\end{equation}
where $\mathcal{L}$ is the log of likelihood in eq.~(\ref{eq:loglikelihood}). The model has tuning parameters $\lambda_{\alpha 0}$ and $\lambda_{\beta 0}$ for enforcing the degree of centrality. 

Since the connections are inferred based on the product of $\bm{\alpha}(t)$ and $\bm{\beta}$ as in~(\ref{eq:logitmodel}), we would have identifiability issues in the two embedding components. To ensure partial identifiability in the embedding spaces, we introduce the regularized terms $\lambda_{\alpha 1}\sum_{j=1}^{M}||\bm{\alpha}_{j}(t)||^{2}_{\mathcal{F}}$ and 
$\lambda_{\beta 1}\sum_{j=1}^{M}||\bm{\beta}_{j}||^{2}$ into the objective function. The first term enforces the $\bm{\alpha}_{j}(t)$'s to be identifiable up to rotations by penalizing the norm, and the second term maintains the identifiability of $\bm{\beta}_{j}$'s. We also use a smoothness penalty $\lambda_{\alpha 2}\sum_{j=1}^{M}||\frac{d\bm{\alpha}_{j}(t)}{dt^{2}}||^{2}_{\mathcal{F}}$ to make sure that embedding functions do not present any volatile changes. The smoothness penalty also spares us in determining the set of basis functions $\phi_{d}(t)$'s since we have more freedom in choosing the number and location of knots. 

The final objective function becomes
\begin{align}
  \mathcal{L}_{\lambda}(\mathbf{\gamma},\mathbf{\beta}) &= \mathcal{L} - \lambda_{\alpha 0}\min_{\{\bm{\omega}_{l}(t)\}_{l=1}^{\text{L}_{\text{out}}}}\sum_{j=1}^{M}\min_{l \in 1, ..., \text{L}_{\text{out}}}||\bm{\alpha}_{j}(t) - \bm{\omega}_{l}(t)||_{\mathcal{F}}^{2} - \lambda_{\beta 0}\min_{\{\bm{\zeta}_{l}\}_{l=1}^{\text{L}_{\text{in}}}}\sum_{j=1}^{M}\min_{l \in 1, ..., \text{L}_{\text{out}}}||\bm{\beta}_{j}- \bm{\zeta}_{l}||^{2}  \nonumber \\ &- \lambda_{\alpha 1}\sum_{j=1}^{M}||\bm{\alpha}_{j}(t)||^{2}_{\Lambda} - \lambda_{\alpha 2}\sum_{j=1}^{M}||\frac{d\bm{\alpha}_{j}(t)}{dt^{2}}||^{2}_{\Lambda} - \lambda_{\beta 1}\sum_{j=1}^{M}||\bm{\beta}_{j}||^{2}.
  \label{eq:finalobj}
\end{align}
The maximization of penalized loglikelihood~(\ref{eq:finalobj}) will be done through gradient updates. Since we cannot simultaneously apply updates on all model components, we perform gradient updates with respect to the embedding components as well as their cluster centers in an alternating fashion. 

Let $h$ represent the iteration index and assume we have initialized the estimates with $\{\bm{\alpha}^{(0)}_{j}(t)\}_{i=1}^{N}$, $\{\bm{\beta}^{(0)}_{j}\}_{j=1}^{N}$, $\{\bm{\omega}^{(0)}_{l}(t)\}_{l=1}^{\text{L}_{\text{out}}}$, and $\{\bm{\zeta}^{(0)}_{l}\}_{l=1}^{\text{L}_{\text{in}}}$. Updating $\bm{\alpha}(t)$ is done through updating the coefficients $\bm{\gamma}$, and we will update $\bm{\gamma}_{j}$ for each node in parallel when the number of nodes $M$ becomes large. That is,
\begin{equation}
  \bm{\gamma}^{(h+1)}_{jr} = \bm{\gamma}^{(h)}_{jr}  + \text{a}_{\alpha} * (\frac{d\mathcal{L}_{\lambda}(\mathbf{\gamma},\mathbf{\beta}^{(h)})}{d\bm{\gamma}^{(h)}_{jr}}) - 2 * \lambda_{\alpha 0 } \min_{l \in 1,..., \text{L}_{\text{out}}} (\bm{\gamma}^{(h)}_{jr} - \bm{\theta}^{(h)}_{lr})^{T}\bm{\Phi} - 2\lambda_{\alpha 1}\bm{\gamma}^{(h)}_{jr}\bm{\Phi} - 2\lambda_{\alpha 2}\bm{\gamma}^{(h)}_{jr}\bm{\Phi}_{2}
  \label{eq:alphaupdate}
\end{equation}
for $r = 1,..., R$ and $\bm{\Phi}_{2}$ is a matrix with $d d^{\prime}$-th entry as $\int \phi^{\prime\prime}_{d}(t)\phi^{\prime\prime}_{d^{\prime}}(t)dt$. 

The update on $\{\bm{\beta}_{j}\}_{j=1}^{M}$ is also done heuristically on each node by applying the gradient update with newly obtained $\{\bm{\alpha}^{(h+1)}_{j}(t)\}_{j=1}^{M}$ involved in $\frac{d\mathcal{L}}{d\bm{\beta}_{j}}$. To obtain $\{\bm{\beta}^{(h+1)}_{j}\}_{j=1}^{M}$, we use the 
\begin{equation}
  \bm{\beta}^{(h+1)}_{j} = \bm{\beta}^{(h)}_{j} + a_{\beta} * \frac{d\mathcal{L}_{\lambda}(\mathbf{\gamma}^{(h+1)},\mathbf{\beta})}{d\bm{\beta}^{h}_{j}} - 2\lambda_{\beta 0}\min_{l\in 1,..., \text{L}_{\text{out}}}(\bm{\beta}^{(h)}_{j}- \bm{\zeta}_{l}) - 2\lambda_{\beta 1}\bm{\beta}^{(h)}_{j}
  \label{eq:betaupdate}
\end{equation}
with learning rates $a_{\alpha}$ and $a_{\beta}$. Given the current embedding components $\{\bm{\alpha}_{j=1}^{(h)}(t)\}_{j=1}^{M}$ and $\{\bm{\beta}_{j}^{(h)}\}_{j=1}^{M}$, the maximization of~(\ref{eq:finalobj}) is equivalent to minimizing the distances of embedding components to cluster centers respectively. Because the distance between $\bm{\alpha}_{j}(t)$ and $\bm{\omega}_{l}(t)$ have been reduced to that between vector coefficients, obtaining both $\bm{\omega}_{l}(t)$ and $\bm{\zeta}_{l}(t)$ can be achieved with standard distance-based clustering algorithms. In this paper, we use the K-mean clustering.

\section{Theoretical Properties}\label{sec:theory}
We establish limiting behaviors and asymptotic distributions of the dynamic/out-going component $\bm{\gamma}_{j}$'s of the proposed embedding method. The asymptotics of the static $\bm{\beta}_{j}$ can be shown similarly while conditioning on the other component. For simplicity notations, we will drop the index for observations and use the generic copy of model components. First, we need the following regularity assumptions:

\begin{itemize}
  \item[(A1)] Embedding functions in all $R$ dimensions are bounded, i.e., there exists a positive constant $C_{\alpha}$ such that $||\alpha_{r}|| < C_{\alpha} < \infty$. 
  \item[(A2)] The penalty parameters are $n$ sequences $\lambda_{\alpha 0n}$, $\lambda_{\alpha 1n}$, and $\lambda_{\alpha 2n}$, and $\lambda_{\cdot n} \rightarrow 0 $ as $n \rightarrow \infty$.
  \item[(A3)] There exists a positive-definite matrix $I$ such chat 
  \begin{equation*}
          \frac{1}{n}\mathbf{\epsilon}^{T}\mathbf{W}\mathbf{\epsilon} + (\lambda_{\alpha 0n} + \lambda_{\alpha 1n})\mathbf{\Phi} + \lambda_{\alpha 2n}\mathbf{\Phi}_{2} \rightarrow I
  \end{equation*}
  as $n \rightarrow \infty$. $\mathbf{\epsilon}$ is a $n$ by $R \times D$ matrix where each row is filled with $\phi_{d}(t_{i})\beta_{kr}$ for $ i = 1,..., n, d = 1,..., D$ , $r = 1, ..., R$ and a fixed $k$. The $\mathbf{W}$ is a $n$ by $n$ diagonal matrix with elements $W_{ii^{\prime}} = \text{var}(P(A_{jk}(t_{i}) = 1))$ for fixed $j$ and $k$. Both $\mathbf{\Phi}$ and $\mathbf{\Phi}_{2}$ are $R \times D$ square matrices with block matrices on the diagonals. The former has elements $\int \phi_{d}\phi_{d^\prime}dt$ to make up a square matrix, whereas the latter has $\int \phi^{\prime\prime}_{d}\phi^{\prime\prime}_{d^\prime}dt$. 
  \item[(A4)] The third partial derivatives of penalized log-likelihood with respect to $\mathbf{\gamma}_{j}$ exist and are finite. 
  \item[(A5)] Let $\tau_{0} = 0 ,...,\tau_{Q} = T$ be the sequence of knots for the spline basis functions $\phi_{d}(t)$'s and assume they are equally spaced, i.e., $\tau_{q} - \tau_{q-1} = h$ for $q = 1,..., D$.
\end{itemize}

\begin{theorem} Given that the regularity assumptions A1-5 are met, then there exists a maximizer $\hat{\mathbf{\gamma}}$ to the penalized likelihood~(\ref{eq:finalobj}) such that $||\hat{\mathbf{\gamma}} - \mathbf{\gamma}_{0}|| = O_{p}(n^{-\frac{1}{2}} + \xi_{n})$ in an open neighborhood of the true coefficients $\mathbf{\gamma}_{0}$ with $\xi_{n} =\max\{\lambda_{\alpha 0n } \min_{l \in 1,..., \text{L}_{\text{out}}} (\bm{\gamma}_{0} - \bm{\theta}_{l})^{T}\bm{\Phi} +\lambda_{\alpha 1n}\bm{\gamma}^{T}_{0}\bm{\Phi} +\lambda_{\alpha 2n}\bm{\gamma}^{T}_{0}\bm{\Phi}_{2}\}$.
  \label{thm:consistency}
\end{theorem}
The coefficients $\mathbf{\gamma}_{0}$ are true in the sense that the $\mathbb{E}(\nabla \mathcal{L}_{\lambda}(\mathbf{\gamma}_{0})) = 0$. The proof is taking some comments and properties of general linear models and likelihood in \cite{scad}, \cite{mleconsistency}, and \cite{GLM}. Some facts and properties of penalized splines have been discussed in \cite{asymp_generalizedsplines}, \cite{localasymptotics}, \cite{localasymptotics2} and \cite{penalizedLR}.

\noindent
{\bf Proof}{ Proving Theorem~\ref{thm:consistency} is equivalent to show that the following probability holds 
\begin{equation*}
  \mathbb{P}(\sup_{||\mathbf{u}|| = C_{u}}|\mathcal{L}_{\lambda}(\gammaT + a_{n}\mathbf{u}) - \mathcal{L}_{\lambda}(\gammaT)|>\delta) < 1 - \epsilon
\end{equation*}
for all $\epsilon > 0 $ and chosen positive $\delta$ and $C_{u}$. Let $a_{n} = n^{-\frac{1}{2}} + \xi_{n}$, and take the Taylor expansion of $\mathcal{L}_{\lambda}(\gammaT + a_{n}\mathbf{u})$ around $\gammaT$, which gives 
\begin{align*}
  \mathcal{L}_{\lambda}(\gammaT + a_{n}\mathbf{u}) &= \mathcal{L}_{\lambda}(\gammaT) + a_{n}\nabla_{\mathbf{\gamma}} \mathcal{L}_{\lambda}(\gammaT)^{T}\mathbf{u} + \frac{1}{2}a_{n}^{2}\mathbf{u}^{T}\nabla^{2}_{\mathbf{\gamma}}\mathcal{L}_{\lambda}(\gammaT)\mathbf{u} \\
  \mathcal{L}_{\lambda}(\gammaT + a_{n}\mathbf{u}) - \mathcal{L}_{\lambda}(\gammaT) &= a_{n} \{ \jacobT - [(\lambda_{\alpha 0Q} + \lambda_{\alpha 1Q})\mathbf{\gammaT} - \mathbf{\theta}_{l}]^{T}\mathbf{\Phi}_{0} - \lambda_{\alpha 2Q}\mathbf{\Phi}_{1}\}\mathbf{u} \\
  &+ \frac{1}{2}a_{n}^{2}\mathbf{u}^{T}\{\nabla^{2}_{\mathbf{\gamma}}\mathcal{L}(\gammaT) + (\lambda_{\alpha 0Q} + \lambda_{\alpha 1Q})\mathbf{\Phi}_{0} + \lambda_{\alpha 2Q}\mathbf{\Phi}_{1}\}\mathbf{u}  
\end{align*}
involving six terms. For the first term, the score function $\jacobT$ goes to the zero on the order of $O_{p}(\sqrt{n})$ which makes the entire term on the order of $O_{p}(a_{n}\sqrt{n})$. The second and third term are on the order of $O_{p}(a_{n}(\lambda_{0Q} + \lambda_{1Q})C_{0}(d+1)h)$ and $O_{p}(a_{n}\lambda_{2Q}C_{0}(d-1)h)$. This is based on the fact that both $\bm{\Phi}$ and $\bm{\Phi}_{2}$ are band matrices filled with inner products of B-spline function. Given assumption A3 and A4, $\frac{1}{2}a_{n}^{2}\mathbf{u}^{T}\nabla^{2}_{\mathbf{\gamma}}\mathcal{L}(\gammaT)\mathbf{u}$ is on the order of $O_{p}(a_{n}^{2}n)$ where $\frac{1}{n}\nabla^{2}_{\mathbf{\gamma}}\mathcal{L}_{\lambda}(\bm{\gamma}) = I(\gammaT)\{1 + o_{p}(1)\}$. The last two terms are on the order of  $O_{p}(a_{n}^{2}(\lambda_{0Q} + \lambda_{1Q})(d+1)h)$ and $O_{p}(a_{n}\lambda_{2Q}(d-1)h)$. The theorem holds by choosing a large enough $C_{u}$.
} \hfill

\begin{remark} 
  Given that the Theorem~(\ref{thm:consistency}) holds, then the estimated coefficients exist and have the following asymptotic distribution $\sqrt{n}(\hat{\mathbf{\gamma}} - \gammaT) \rightarrow \text{N}(0, F(\lambda)^{-1}F_{\lambda=0}F(\lambda)^{-1})$ where $F_{\lambda} = \mathbf{\epsilon}^{T}\mathbf{W}\mathbf{\epsilon} + (\lambda_{\alpha 0} + \lambda_{\alpha 1})\mathbf{\Phi} + \lambda_{\alpha 2}\mathbf{\Phi}_{2}$ 
\label{rmk:asynormal}
\end{remark}

\section{Simulation Studies}
\label{sec:simulation}
In the simulation studies, we compare the proposed representation learning model with LINE \citep{LINE}, Deepwalk \citep{deepwalk}, and Dynnode \citep{dynnode2vec}. First, we start to simulate the dynamic network with fixed parameters $R = 6$ the embedding dimension, $\Lout = 4$ the number of clusters for $\bm{\alpha}(t)$, and $\Lin = 5$ the number of clusters for $\bm{\beta}$. Let $\bm{\mu}^{\gamma}_{l}$ to be cluster center for $l$-th out-going embedding components in terms of the coefficients, then individual coefficients $\bm{\gamma}_{ir}$ are generated from a multivariate normal $\text{N}(\bm{\mu}^{\gamma}_{l}, \sigma^{2}_{\alpha}\bm{\text{I}})$ for $r = 1, ..., 6$. The noise parameter $\sigma^{2}_{\alpha}$ controls the variability in coefficients hence the out-going embedding functions within a cluster. Similarly, the receiving embedding vectors $\bm{\beta}_{j}$ are generated from a multivariate normal $\text{N}(\bm{\mu}^{\beta}_{l}, \sigma^{2}_{\beta}\bm{\text{I}})$ given $l = 1, ..., \Lin$. Given a set of pre-determined basis functions $\{\phi_{d}(t)\}$'s, we can obtain the embedding components $\bm{\alpha}_{j}(t)$'s based on the simulated coefficients. As a result, the longitudinal logits at a sequence of points $t_{1}, ..., t_{n}$ can be calculated as 
\begin{equation*}
  \eta_{jk}(t_{i}) = \langle \bm{\alpha}_{j}(t_{i}), \bm{\beta}_{k} \rangle
\end{equation*}
which gives the probability of generating a connection from $j$-th node to $k$-th node at time point $t_{i}$ for $i = 1, ..., nQ$.

The first model comparison is on the link prediction of time-varying networks. Let $\{\bm{A}(t_{i})\}_{i=1}^{n}$ be the sequence of time snapshots of the dynamic network. For each snapshot $\bm{A}(t_{i})$, we randomly withhold a portion of connections and use the remaining connections to perform representation learning. After obtaining the outgoing embedding functions and receiving embedding vectors, we then perform link predictions for nodes if any of their connections have been removed at any time point. Table~(\ref{tab:sim1static}) summarizes the F1-score of the link prediction performances of the proposed method and other comparing ones. We calculate the F1-score of link prediction after averaging over $t_{1}$,..., $t_{n}$. The data simulation, model estimation, and prediction are repeated 100 times to assess the consistency of these methods.

\begin{table}[htbp]
  \centering
  
  \caption{F1-score of link predictions for the proposed method, LINE, Dynnode, and Deepwalk. Each row corresponds to one method. The percentages represent the portion of links removed from the actual adjacency matrix at each time point. The mean of the F1-score of 100 repetitions is reported along with the standard deviation in the parentheses. \label{tab:sim1static}}\vspace{0.5cm}
  \begin{tabular}{c|c|c|c|c|c}
    \hline 
    & 10\% & 20\% & 30\% & 40\% & 50\%  \\
\hline    
Proposed & 0.919(.026) & 0.915(.025)& 0.911(.026)& 0.904(.029) & 0.896(.03)  \\
\hline
LINE & 0.669(.024)& 0.668(.018)& 0.668(.016)& 0.667(.014)& 0.667(.012) \\
\hline
Dynnode & 0.668(.022)& 0.668(.017)& 0.669(.014)& 0.669(.013)& 0.669(.012) \\
\hline
Deepwalk & 0.663(.022)& 0.663(.015)& 0.663(.012)& 0.662(.01)& 0.662(.009) \\
\hline

\end{tabular}
\end{table}

Table~(\ref{tab:sim1static}) shows that the proposed method is superior in interpolating dynamic networks where the structure is not fully observed. The performance of link prediction is consistent regardless of the number of missing links. In addition to predicting the missing links within a time point, we can also assess whether different learning methods can recover entire unobserved network structures at time points. The following simulation considers withholding network information at some time points and uses the remaining networks to perform model estimation. 

\begin{figure}[htbp]
  \centering
  \includegraphics[width=\textwidth]{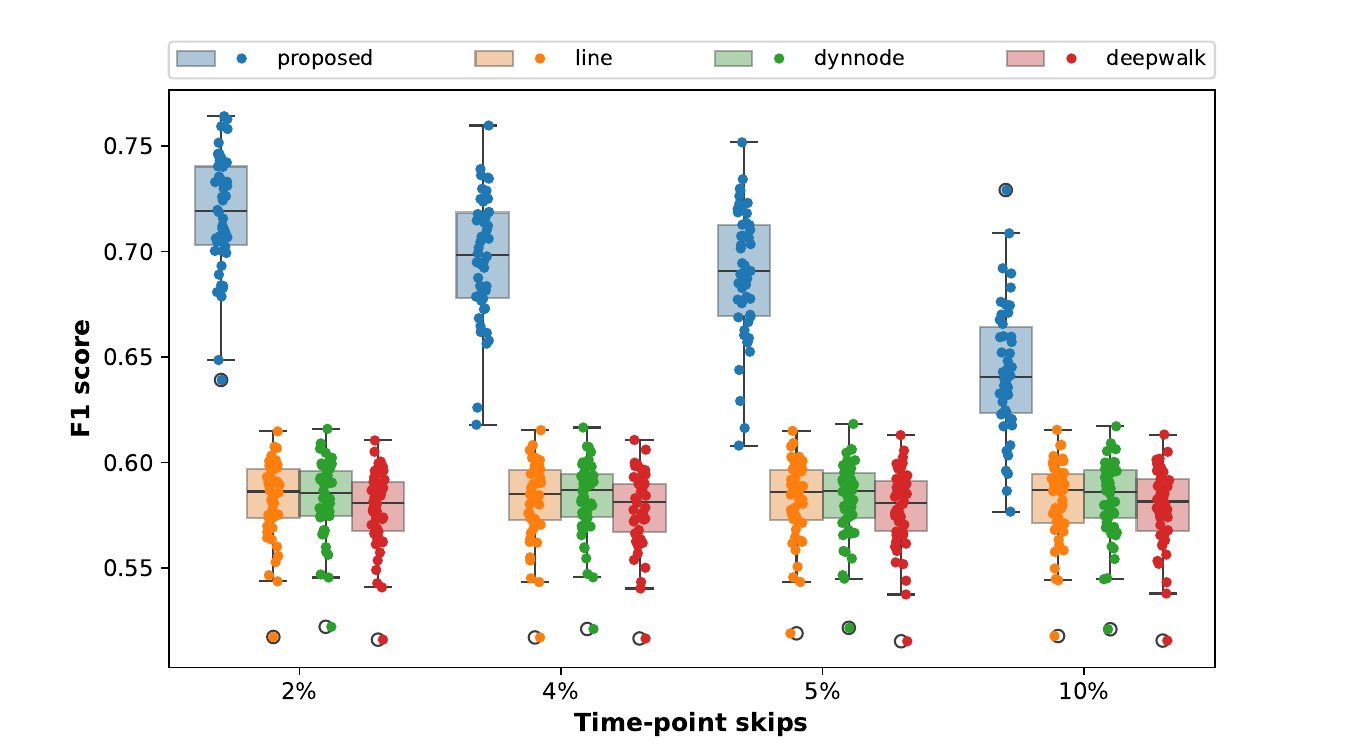}
  
  \caption{The percentages represent the portion of time points that are skipped for estimating the time-dependent node embeddings. For example, $2\%$ represents that $2\%$ of $n$ time points have been skipped between a pair of actually observed adjacency matrices.\label{fig:f1scoreavgtime_dynamic_sim1}}
\end{figure}

Figure~\ref{fig:f1scoreavgtime_dynamic_sim1} summarizes the link prediction of such simulation which has been repeated 50 times. Figure~\ref{fig:f1scoreavgtime_dynamic_sim1} shows that the proposed method is not only good at recovering partially observed networks within a time point but also can recover an entire network structure at time points where we do not have observations. Additional simulation results and computing codes are included in the supplementary materials.

\section{Analysis on the Dynamic Network of Ant Colonies}
\label{sec:applications}
\begin{figure}[htbp]
  \centering
  \includegraphics[width=\textwidth]{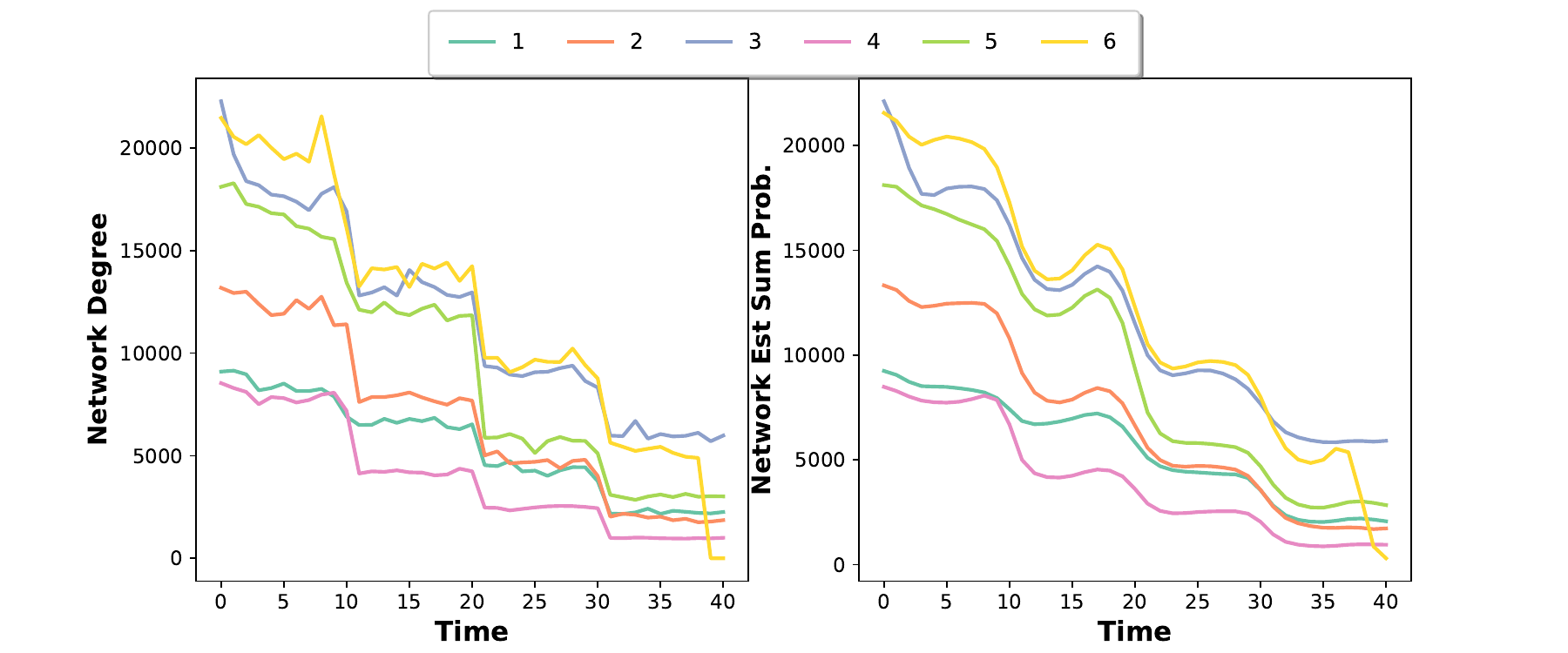}
  
  \caption{On the left: The total degree of the dynamic network structures ants in each colony over time. On the right: the sum of estimated probabilities of connections at each time point. The color indicates the ant colony.\label{fig:ant_degreeprob}}
\end{figure}

The application is carried out on a data set provided in \cite{antnet} which was used to study the sociological networks of ant colonies. The data set is composed of 6 dynamic networks from 6 ant colonies. Each network has a different ant population, but all networks are observed on the 41-day time span. For each day, we summarize the social interactions of the ant population in an adjacency matrix. From \cite{antnet}, each ant population is composed of roughly 4 groups: the queen, nursers, cleaners, and foragers. The ants will take responsibilities from nursers to cleaners to foragers as they grow and be active at a larger radius from the queen. Figure~(\ref{fig:ant_degreeprob}) depicts network activities as well as the estimated connectivities over time. 
On the left of Figure~\ref{fig:ant_degreeprob}, we can see the decrease in the network degree (total number of connections in a network) over time. The observed degrees over time are very rough while the estimated version on the right gives much smoother trajectories. This is the benefit of using functional data analysis which allows us to infer network structures at points where we do not have observations. This helps us to explain the network dynamics from point to point.

\begin{figure}[htbp]
  \centering 
  \includegraphics[width=\textwidth]{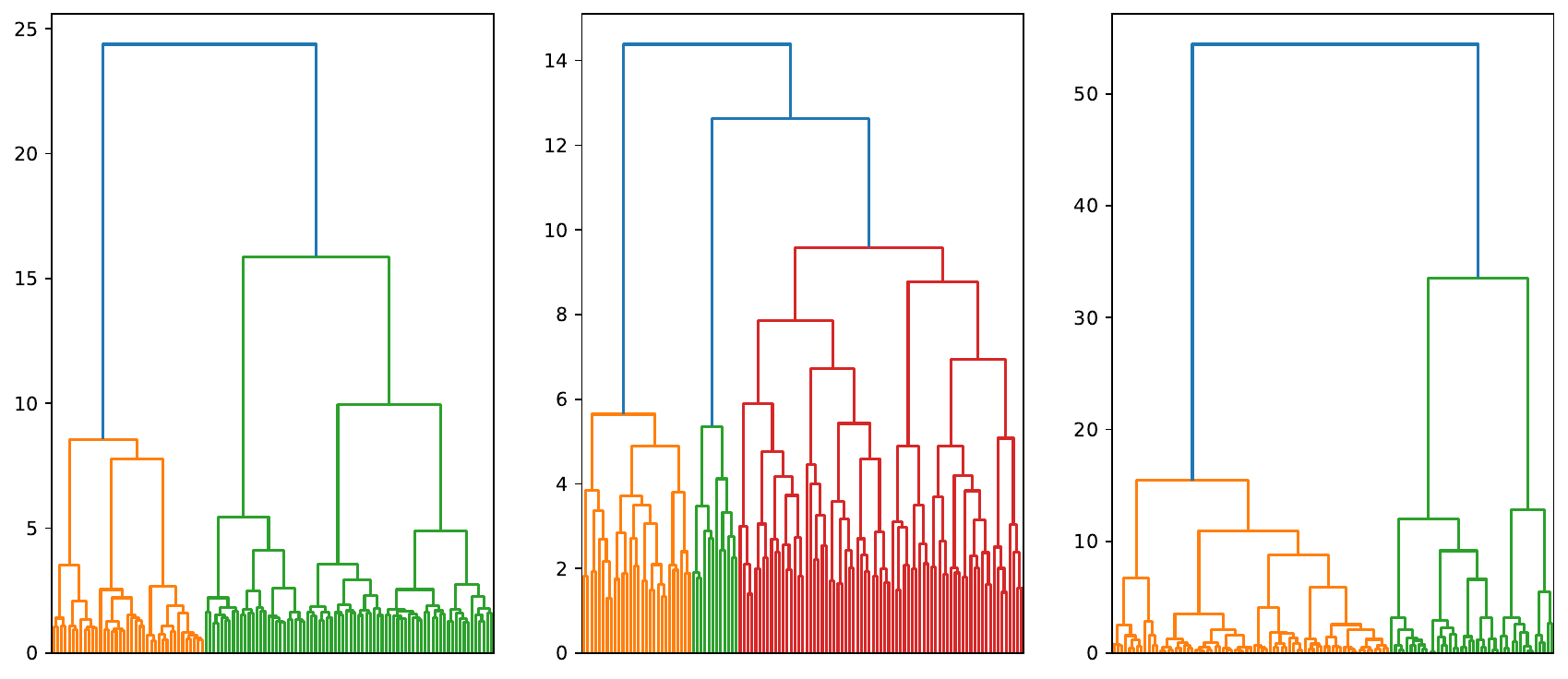}
  \caption{Static clustering of an ant colony. On the left: hierarchical clustering of the network summed over time. In the middle: clustering on the $\bm{\alpha}_{j}(t)$'s. On the left: clustering of the $\bm{\beta}_{j}$'s. \label{fig:ant_net1_hcluster}}
\end{figure}
Figure~\ref{fig:ant_net1_hcluster} demonstrates the hierarchical clustering results of the network. The left demonstrates the hierarchical clustering results when we sum the adjacency matrices over time. The number of clusters in the leftmost plot in Figure~\ref{fig:ant_net1_hcluster} is approximately consistent with the prior information about 4 social groups in the network. In the middle of Figure~\ref{fig:ant_net1_hcluster}, we have the clustering based on the out-going embedding function $\bm{\alpha}_{j}(t)$ of individual ants, and it is done by doing hierarchical clustering on the coefficients $\bm{\gamma}_{i}$'s. The clustering and community structure is more complex of the out-going embedding (in the middle plot of Figure~\ref{fig:ant_net1_hcluster}) compared to that of the receiving embedding (in the right-most plot of in Figure~\ref{fig:ant_net1_hcluster}). The proposed method can provide asymmetric embeddings of individual nodes and give a more comprehensive understanding of network dynamics in comparison to other embedding methods.

\begin{figure}[htbp]
  \centering
  \includegraphics[width=\textwidth]{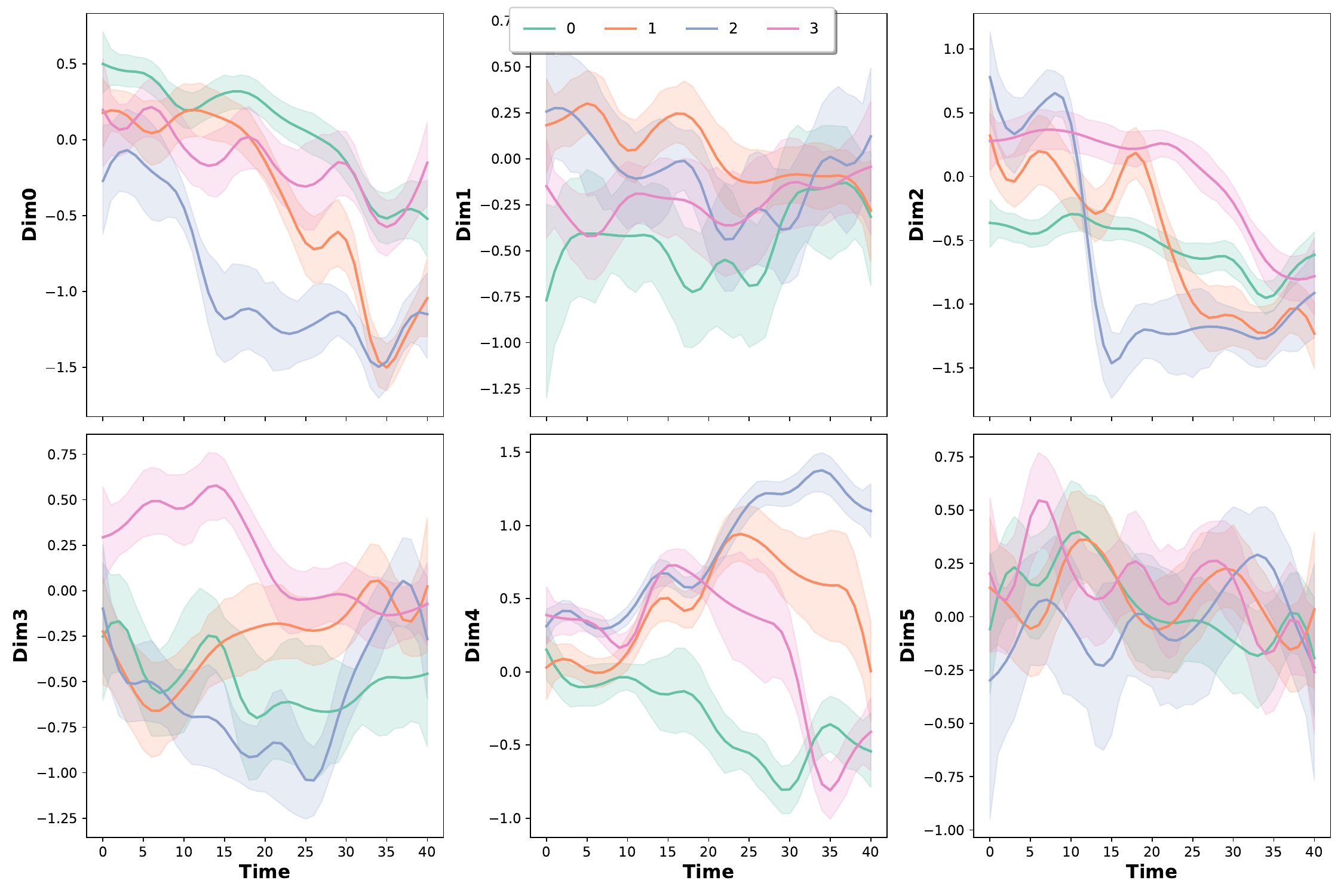}
  \caption{Display of the out-going embedding function $\bm{\alpha}_{j}(t)$'s. Different colors indicate different clusters of $\bm{\alpha}_{j}(t)$'s. The solid line represents the pointwise mean and the shaded area demonstrates the values of the embedding functions. \label{fig:ant_net1_embout}}
\end{figure}

\begin{figure}[htbp]
  \centering
  \includegraphics[width=\textwidth]{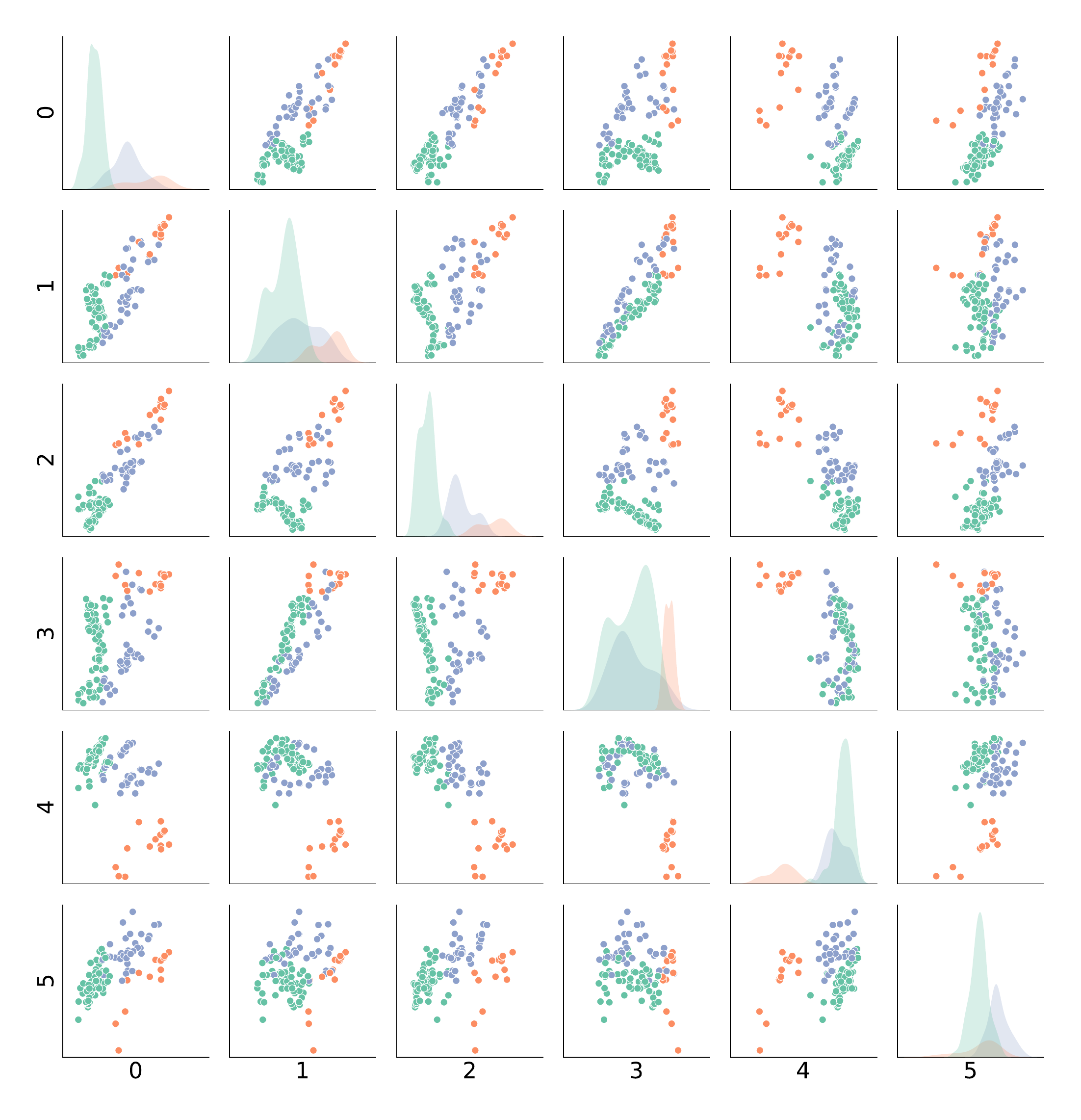}
  \caption{Scatter plot matrix of $\bm{\beta}_{j}$'s. Each color indicates a different cluster. \label{fig:ant_net1_embin}}
\end{figure}

We can also take a closer look at the out-going embedding function $\bm{\alpha}_{j}(t)$'s which are summarized in Figure~\ref{fig:ant_net1_embout}.
We have chosen $R = 6$ to be the embedding dimension, we can see that some dimensions (Dim 0,2,4 in Figure~\ref{fig:ant_net1_embout}) are more significant in distinguishing different clusters. This clustering of $\bm{\alpha}_{j}(t)$'s is done statically through the coefficients. The choice of 4 clusters is based on the prior information on the social structures of ants. We can also check the clustering of the receiving embedding information in $\bm{\beta}_{j}$'s as shown in Figure~\ref{fig:ant_net1_embin}.

\begin{figure}[htbp]
  \centering
  \includegraphics[width=\textwidth]{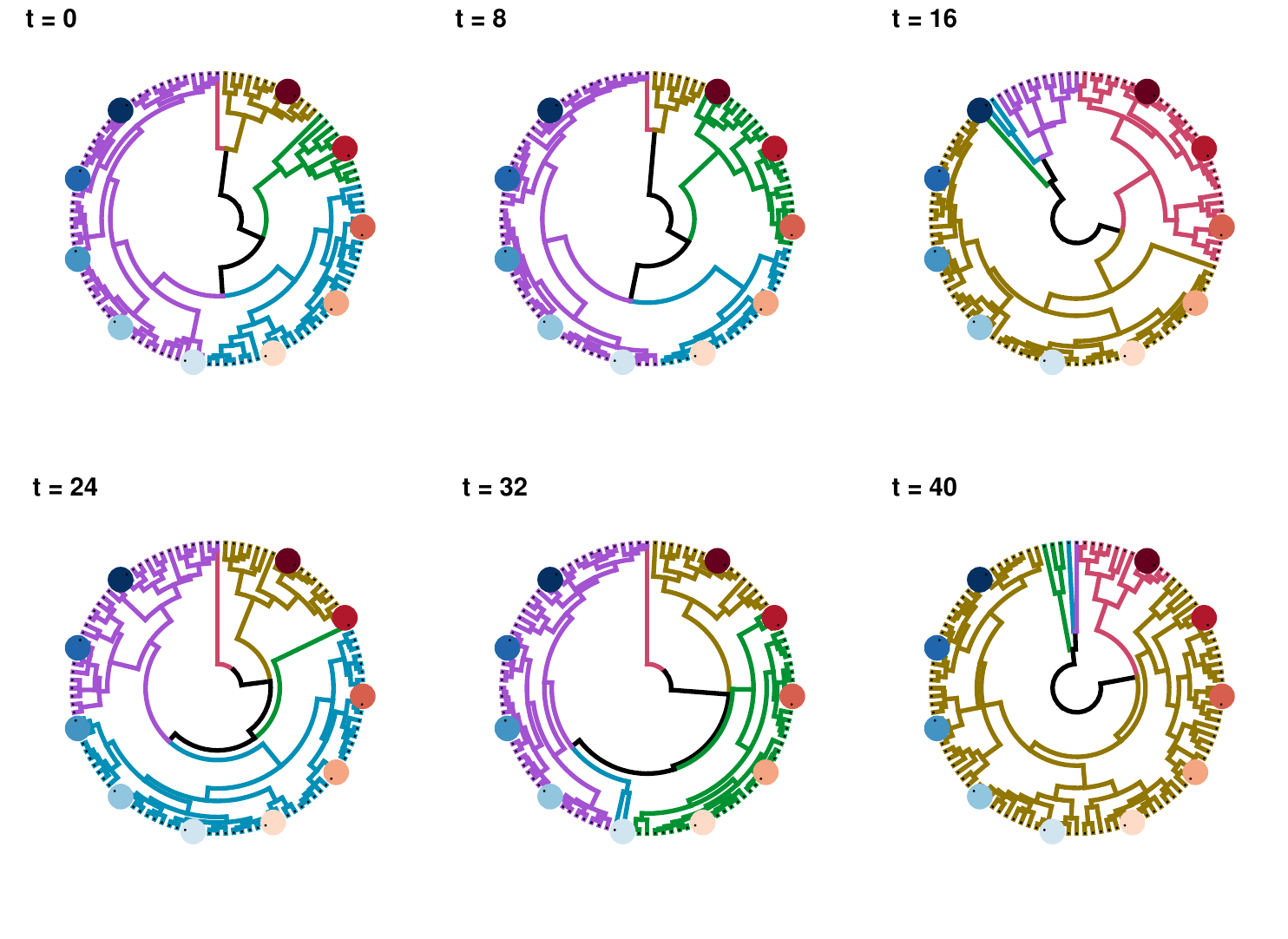}
  \caption{Dynamic clustering of the out-going embedding function at $t = 0, 8, 16, 24, 32, 40$. \label{fig:ant_net1_timedependentcluster}}
\end{figure}

The clustering of the receiving embedding $\bm{\beta}_{j}$'s is more obvious than that of the outgoing embeddings. Previously, we have displayed and compared the static clustering of the embedding function $\bm{\alpha}_{j}(t)$. Since these embedding functions are in a metric space, we can always calculate pairwise distances between node embeddings at a certain time point. Hence, we can do dynamic clustering of $\bm{\alpha}_{j}(t)$ along $t$, and the result is illustrated in Figure~\ref{fig:ant_net1_timedependentcluster}.
The dynamic clustering of $\bm{\alpha}_{j}(t)$ is consistent with both prior knowledge about an ant society and initial assessment of the network dynamic as in Figure~(\ref{fig:ant_degreeprob}). At time $t = 0$, the ants are relatively close to each other, hence we see that the clusters are roughly equal in size. As $t$ increases and ants grow older, we see a cluster has its size increasing. This includes ants that are older and those who become foragers of the colony. They got more separated from the Queen and new ones since they were working and remaining on the far perimeter of the colony. In the meantime, other clusters have shrunk in size since the colony population was fixed during the experiments.

Furthermore, we have assessed the consistency of our proposed method and compared it with the existing approaches. Similar to the simulation studies, we want to see whether the proposed method can better predict missing links of the network across time points. Based on Table~\ref{tab:antstatic}, our proposed method is superior in predicting missing links in the network. 
\begin{table}[htbp]
  \centering
  \caption{F1-score of link predictions for the proposed method, LINE, Dynnode, and Deepwalk. Each row corresponds to one method. The percentages represent the portion of links removed from the actual adjacency matrix at each time point. The mean of the F1-score of 50 repetitions is reported along with the standard deviation in the parentheses. \label{tab:antstatic}}\vspace{0.5cm}
  \begin{tabular}{c|c|c|c|c|c}
    \hline 
    & 10\% & 20\% & 30\% & 40\% & 50\%  \\
\hline    
Proposed &0.956(.0042)& 0.951(.0045)& 0.947(.05)& 0.944(.051)& 0.933(.062) \\
\hline 
LINE & 0.812(.0091)& 0.813(.009)& 0.811(.089)& 0.808(.089)& 0.802(.088) \\
\hline 
Dynnode & 0.657(.023)& 0.658(.016)& 0.657(.013)& 0.658(0.012)& 0.657(0.01) \\
\hline 
Deepwalk & 0.825(.108)& 0.807(.119)& 0.802(.123)& 0.799(.125)& 0.796(.127) \\
\hline
\end{tabular}
\end{table}

\section{Conclusions and Discussion}
\label{sec:discussion}

In this work, we propose a representation learning method for dynamic networks. The method provides asymmetric embeddings of individual nodes in comparison to the existing approaches. Furthermore, the most significant contribution is that our embeddings are dynamic rather than static or calculated discretely on time points. This allows us to infer network structures at any time point with or without actual observations of the network. Given the benefits of using functional data analysis, the embedding functions are endowed with metrics to easily calculate distances between dynamic embeddings or embeddings at a certain time point. The proposed method produces embeddings that are much more useful in predicting network structure given partially observed networks or networks observed on sparse time points. 

Further works will consider the scalability of the method to large or long networks. Also, we can explore how to use extrinsic information (other covariates observed for individual nodes in the network) to obtain embeddings as well as build classification/regression models based on node embeddings. 


\vskip 0.2in
\bibliographystyle{apalike}
\bibliography{dne_bib}

\end{document}